\documentclass[10pt,twocolumn,letterpaper]{article}

\usepackage{cvpr}              

\usepackage[accsupp]{axessibility}  

\newcommand{\blue}[1]{{\color{blue}#1}}

\newcommand{\model}{$x^2$-Fusion}

\makeatletter
\renewcommand\thanks[1]{%
  \footnotemark[0]%
  \protected@xdef\@thanks{\@thanks%
    \protect\footnotetext[0]{\setlength{\parindent}{0pt}#1}%
  }%
}
\makeatother

\definecolor{cvprblue}{rgb}{0.21,0.49,0.74}
\usepackage[pagebackref,breaklinks,colorlinks,allcolors=cvprblue]{hyperref}
\usepackage{booktabs} 
\usepackage{multirow} 
\usepackage{array}    
\usepackage[table]{xcolor}
\usepackage{amsmath}
\usepackage{amssymb}
\usepackage{algorithm}
\usepackage{algorithmic}

\def\paperID{2887} 
\def\confName{CVPR}
\def\confYear{2026}

\title{$x^2$-Fusion: Cross-Modality and Cross-Dimension Flow Estimation \\in Event Edge Space}

\makeatletter
\renewcommand\thanks[1]{%
  \footnotemark[0]%
  \protected@xdef\@thanks{\@thanks%
    \protect\footnotetext[0]{\setlength{\parindent}{0pt}#1}%
  }%
}
\makeatother

\author{
    Ruishan Guo$^{1,*}$,
    Ciyu Ruan$^{1,*}$,
    Haoyang Wang$^{1,*}$,
    Zihang Gong$^{2}$,
    Jingao Xu$^{3}$,
    Xinlei Chen$^{1,\dagger}$
    \thanks{%
    \textsuperscript{*}Equal Contribution.
    \textsuperscript{\dag}Corresponding author.%
    }\\
    $^{1}$Shenzhen International Graduate School, Tsinghua University,
    $^{2}$Harbin Institute of Technology,\\
    $^{3}$The University of Hong Kong
    \\
    {\tt\small \{grs24, rcy23, haoyang-22\}@mails.tsinghua.edu.cn,
    gongzihang0201@gmail.com}
    \\
    {\tt\small jingaoxu@hku.hk,
    chen.xinlei@sz.tsinghua.edu.cn}
}

\begin{document}
\maketitle
\begin{abstract}

Estimating dense 2D optical flow and 3D scene flow is essential for dynamic scene understanding. Recent work combines images, LiDAR, and event data to jointly predict 2D and 3D motion, yet most approaches operate in separate heterogeneous feature spaces. Without a shared latent space that all modalities can align to, these systems rely on multiple modality-specific blocks, leaving cross-sensor mismatches unresolved and making fusion unnecessarily complex.
Event cameras naturally provide a spatiotemporal edge signal, which we can treat as an intrinsic edge field to anchor a unified latent representation, termed the \textbf{Event Edge Space}. Building on this idea, we introduce \textbf{\model}, which reframes multimodal fusion as representation unification: event-derived spatiotemporal edges define an edge-centric homogeneous space, and image and LiDAR features are explicitly aligned in this shared representation.
Within this space, we perform reliability-aware adaptive fusion to estimate modality reliability and emphasize stable cues under degradation. We further employ cross-dimension contrast learning to tightly couple 2D optical flow with 3D scene flow. 
Extensive experiments on both synthetic and real benchmarks show that \model\ achieves state-of-the-art accuracy under standard conditions and delivers substantial improvements in challenging scenarios.

\end{abstract}
\vspace{-0.5cm}
\section{Introduction}

Optical and scene flow estimate dense 2D and 3D motion correspondences across images, depth maps, and point clouds, forming a core tool for dynamic scene understanding in autonomous driving, tracking, and 3D reconstruction~\cite{dosovitskiy2015flownet, wang2025ultra, wang2020flownet3d++}. 
Recent works fuse image, LiDAR, and events via advanced architectures, coupling dense photometry with accurate geometry to exploit complementarity and outperform single-modality baselines~\cite{huang2022flowformer,xu2022gmflow,luo2025mambaflow, ruan2025pre}.

\begin{figure}[h]
    \centering
        \includegraphics[width=1\linewidth]{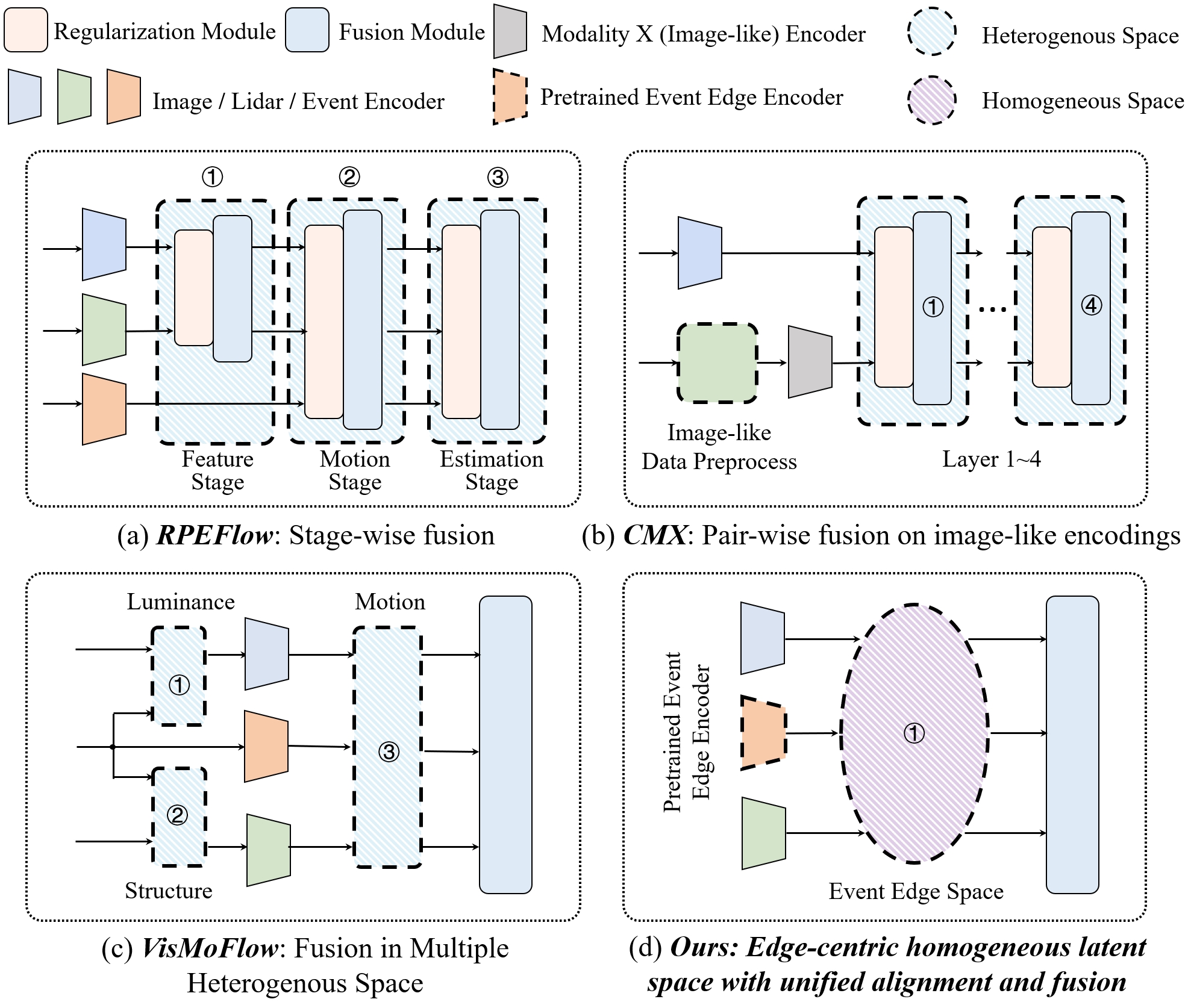}
    \vspace{-0.6cm}
    \caption{\textbf{Fusion paradigms for multimodal perception.}
(a) RPEFlow~\cite{wan2023rpeflow}: stage-wise fusion across 2D/3D spaces.
(b) CMX~\cite{zhang2023cmx}:image-like encodings with pairwise rectification/attention at each backbone stage.
(c) VisMoFlow~\cite{zhou2024bring}: separate luminance/structure/correlation spaces with dedicated modules.
(d) Ours: preserves native domains and learns an \emph{Event Edge Space}, a shared edge-centric latent space guided by a frozen event teacher, that aligns image, LiDAR, and events before fusion.}
    \label{fig:paradigms}
    \vspace{-0.7cm}
\end{figure} 

Despite these advances, existing multimodal flow estimation works keep each modality in its native format (images as 2D grids, LiDAR as point clouds, events as asynchronous streams) and fuse them spatially across separate, \emph{heterogeneous} feature spaces.
This brings three issues: 
\\ \emph{(i)  High Complexity}: Without a shared channel-wise basis, fusion requires pairwise alignment between every modality pair, leading to module-heavy architectures with stage-wise fusion blocks~\cite{wan2023rpeflow}, pair-wise rectification/attention units~\cite{zhang2023cmx}, or multiple hand-crafted physical spaces~\cite{zhou2024bring} (Fig.~\ref{fig:paradigms}(a-c)). This makes models cumbersome, difficult to train, and hard to scale to additional modalities.
\\ \emph{(ii) Information Erosion}: Processing features in separate heterogeneous spaces delays fusion to late stages, where signals have already been corrupted by early-stage modality-specific distortions that become difficult to correct through cross-modal interaction.
\\ \noindent\emph{(iii) High Fragility}: Without a common representational foundation, modalities cannot provide stable priors for one another. Under perception degradations such as exposure extremes, LiDAR sparsity, or motion blur, the alignment itself breaks down, causing catastrophic failures~\cite{wang2026mme, kong2023robodepth, luo2024eventtracker, wei2022lidar}.
Overall, these methods do not align all sensors into a \emph{homogeneous} channel-wise latent space, preventing simple, robust, and efficient cross-modal interaction.
\noindent\textbf{Multimodel Fusion in Homogeneous Event Edge Space.}
We address this limitation by introducing \textbf{\emph{Event Edge Space}}, the first homogeneous latent space that unifies image, LiDAR, and event representations in shared edge-centric domain. This design is motivated by two insights:

\noindent $\bullet$ \textit{Why Edge?} 
Edges capture object boundaries and scene discontinuities in a modality-agnostic manner. They represent consistent structural information across sensors, irrespective of appearance variations, sampling density differences, or sensor-specific noise patterns~\cite{zhao2024edge,liu2024event,ruan2025edmamba,zhou2022edge}. An edge-centric latent space thus provides a natural common language for aligning heterogeneous inputs before interaction.

\noindent $\bullet$ \textit{Why Edge using Event?} 
Event camera records per-pixel brightness changes at ultra-high temporal resolution, firing precisely where strong image gradients persist under motion, i.e., on moving edges~\cite{gallego2020event, wang2026event}. Spatial aggregation traces pixel-aligned edge curves; temporal aggregation yields continuous edge trajectories, forming a natural spatiotemporal edge signal. 
Moreover, events share 2D pixel coordinates with images, while their asynchronous sparse activations mirror LiDAR's irregular spatiotemporal sampling~\cite{li2024sparsefusion,wang2025towards}.
This dual correspondence—geometric alignment with image and structural similarity with LiDAR—makes events an ideal anchor for a truly homogeneous edge-centric space.



To translate this insight into a practical architecture, we introduce three key technical components. 
First, we pretrain an \textbf{\emph{Event Edge Encoder}} to distill stable, motion-aware edge embeddings, then freeze it and use its features as edge prototypes to \emph{symmetrically regularize} RGB and LiDAR encoders, aligning both modalities into the shared Event Edge Space. 
Second, within this homogeneous space, \textbf{\emph{Reliability-aware Adaptive Fusion}} estimates global and local reliability from spatiotemporal cues and performs cross-attention to produce unified 2D/3D features.
Third, \textbf{\emph{Cross-dimension Contrastive Learning}} explicitly enforces inter-frame motion coherence and 2D–3D geometric consistency, enabling mutual reinforcement between optical flow and scene flow tasks to fully exploit complementary cues.
Extensive experiments on both synthetic and real-world benchmarks demonstrate that our approach achieves state-of-the-art performance under normal conditions. Moreover, it significantly outperforms existing methods under challenging scenarios such as extreme lighting and LiDAR sparsity. Our main contributions are summarized as follows:
\begin{itemize}
\item To our knowledge, we introduce Event Edge Space, the first edge-centric homogeneous space unifying image, LiDAR, and events representations in a common domain.

\item Within the homogeneous space, we couple reliability-aware adaptive fusion with cross-dimension contrastive learning, enabling consistent 2D-3D flow estimation.
\item We achieve state-of-the-art performance on both synthetic and real-world benchmarks, demonstrating the effectiveness of our approach across diverse scenarios.
\end{itemize}

\vspace{-0.2cm}
\section{Related Work}
\subsection{Multimodel Fusion Paradigms}
Multimodal sensing enhances scene understanding by combining dense appearance, precise geometry, and fine-grained dynamics~\cite{liu2022camliflow, li2022deepfusion, guo2025unsupervised, wang2025enabling, wan2025instance, ding2025hawkeye}. In recent 2D/3D motion tasks, coupling photometric cues with 3D structure~\cite{teed2020raft, wan2023rpeflow, zhang2023cmx, zhou2024bring} significantly outperforms single-modality baselines~\cite{liu2019flownet3d, gu2019hplflownet, li2021neural, zhang2025ematch, luo2025learning, raju2025perturbed}.

However, most prior work models pairwise interactions across heterogeneous feature spaces via bespoke fusion blocks: RPEFlow~\cite{wan2023rpeflow} adopts stage-wise attention and mutual-information regularization at the feature, motion, and estimation stages, requiring dedicated fusion for each branch and stage. 
VisMoFlow~\cite{zhou2024bring} attempts to mitigate modality heterogeneity by constructing three hand-crafted physical spaces:luminance, structure, correlation. Although these are described as \textit{homogeneous spaces}, the three domains are processed by distinct modules (luminance fusion, structural cross-attention, correlation alignment) and rely on different physical assumptions. Consequently, interaction occurs sequentially across these domains rather than within a unified channel-wise latent space, and the overall fusion mechanism remains multi-stage and module-intensive.
Unified RGB–X methods like CMX~\cite{zhang2023cmx} convert all inputs to image-like encodings and fuse via shared blocks.
While providing a common 2D representation, this grid-based confinement overlooks modality-native properties such as 3D topology and asynchronous event dynamics.

In contrast, we embed images, LiDAR, and events into a homogeneous edge-centric space. This enables fusion via lightweight weighting and unified cross-attention, replacing stacked modality-wise modules to a simpler, robust design.

\begin{figure*}[t]
    \centering
        \includegraphics[width=1.0\linewidth]{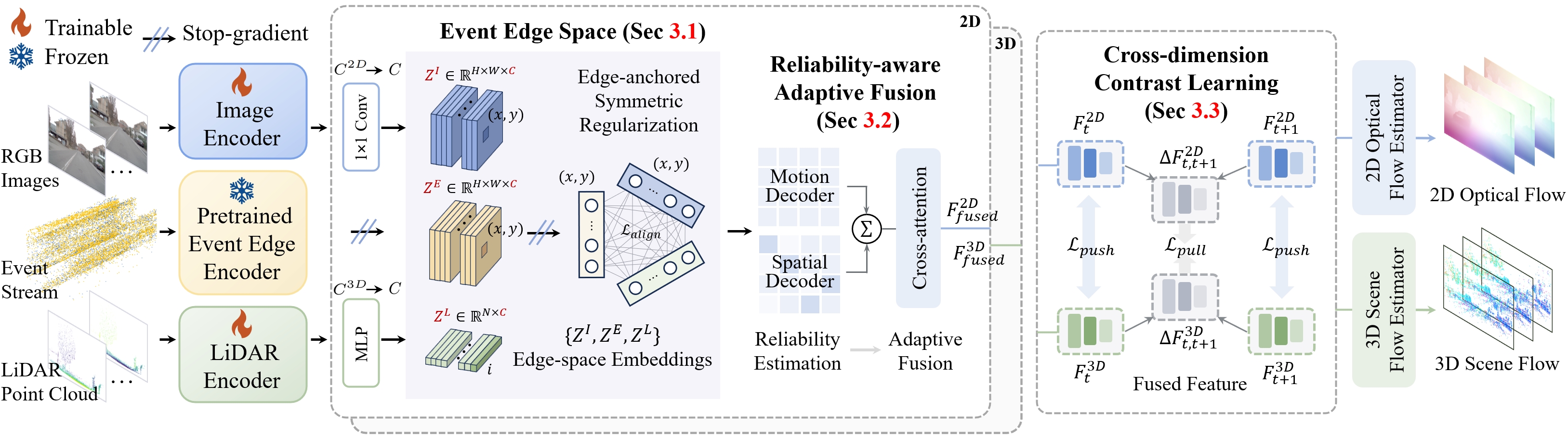}
     \caption{\textbf{Overview of \model.} 
     Given image, events and LiDAR, we first pretrain an \emph{Event Edge Encoder} to distill motion-aware edge features. We then freeze the encoder and use its embeddings as edge prototypes to \emph{symmetrically regularize} channel-wise representations across modalities, aligning them into a shared \textbf{\emph{Event Edge Space}} (Sec.~\ref{sec:event_edge_space}). Within this space, \textbf{\emph{Reliability-aware Adaptive Fusion}} estimates global and local reliability and 
     fuses modalities via a cross-attention block to produce 2D/3D features
     (Sec.~\ref{sec:event_aware_adaption_fusion}). Finally, \textbf{\emph{Cross-dimension Contrast Learning}} enforces inter-frame coherence and 2D–3D consistency, and the task heads output optical and scene flow (Sec.~\ref{sec:CCL}).}
     
    \label{pipeline}
    \vspace{-0.3cm}
\end{figure*} 

\subsection{Cross-modality and Cross-dimension Learning}
Existing cross-modality learning methods primarily focus on: 1) modeling inter-modal relationships~\cite{cui2021rosita, chen2024disentangled, yan2023cross, hussen2020modality, li2023bevdepth} and 2) designing effective fusion strategies~\cite{zhang2025point, hori2017attention, wei2020multi, dosovitskiy2020image, hu2017learning, belghazi2018mutual, zhou2024pointcmc, lyu2022finite, hjelm2018learning}.

For relationship modeling, co-learning approaches such as cross-modal distillation~\cite{yan2023cross} learn shared representations but are prone to failure propagation when any sensor degrades. Modality dropout~\cite{hussen2020modality} improves robustness by simulating sensor failures in training. Knowledge-guided methods inject domain priors (e.g., geometric constraints~\cite{li2023bevdepth}) but often lack adaptability to unseen scenarios.

Fusion strategies generally fall into two categories: attention-based approaches~\cite{hori2017attention, wei2020multi}, which dynamically weight modalities via gating or cross-attention ~\cite{dosovitskiy2020image}, and representation-learning methods~\cite{hu2017learning, belghazi2018mutual}, which align feature spaces by enforcing embedding consistency through contrastive objectives\cite{zhou2024pointcmc}, canonical correlation\cite{lyu2022finite}, or mutual information regularization~\cite{hjelm2018learning}.

We propose a unified fusion framework on a shared, edge-centric latent space. Within this space, cross-modality adaptive attention and cross-dimension contrastive learning combine to enable robust multimodal integration.

\section{Method}

\paragraph{Overview.} 
To enable robust optical and scene flow estimation under all-day, dynamic conditions, we propose a unified framework named \textbf{\model}, as illustrated in Fig.~\ref{pipeline}, that uses stable scene-edge representations to unify image, events and LiDAR in a shared edge-centric latent space, and then performs reliability-aware adaptive fusion with joint 2D--3D motion representation learning.

For robust multimodal fusion, we first embed image, LiDAR, and events into a homogeneous \textbf{\emph{Event Edge Space}}, where frozen event embeddings provide edge-centric prototypes to align modality features. Within this shared space, \textbf{\emph{Reliability-aware Adaptive Fusion}} estimates modality reliability from stable edge motion and structural cues and drives a single cross-attention block, yielding resilient 2D/3D representations under partial sensor degradations.

For joint estimation, the \textbf{\textit{Cross-dimension Contrast Learning}} module jointly optimizes 2D and 3D motion representations by enforcing intra-frame geometric discrimination and inter-frame consistency, enabling cross-dimension synergy that surpasses independent learning.

\subsection{Event Edge Space}
\label{sec:event_edge_space}
We propose the \textbf{\textit{Event Edge Space}} to unify multimodal features based on the intrinsic geometric commonality of motion edges across all three modalities. Leveraging the fact that event streams are inherently composed of fine-grained spatiotemporal edge points, we exploit their temporally dense and edge-salient nature to construct this homogeneous latent space. It serves as a unified feature domain, anchored by the semantics of motion edges.

To faithfully preserve fine-grained spatiotemporal edge cues within this space, we first explicitly pretrain an event edge encoder (Sec.~\ref{sec:event_edge_encoder}) to distill stable, motion-aware edge representations from raw events. We then freeze the event encoder, treating its features as Edge Prototypes to learn bidirectional mappings $\text{I}\Leftrightarrow \text{E}$ and $\text{L}\Leftrightarrow \text{E}$ for aligning image and LiDAR features (Sec.~\ref{sec:RGB--LiDAR Alignment in Event Edge Space}).

\subsubsection{Event Edge Encoder Pretraining}
\label{sec:event_edge_encoder}

\begin{figure}[t]
    \centering
        \includegraphics[width=1.0\linewidth]{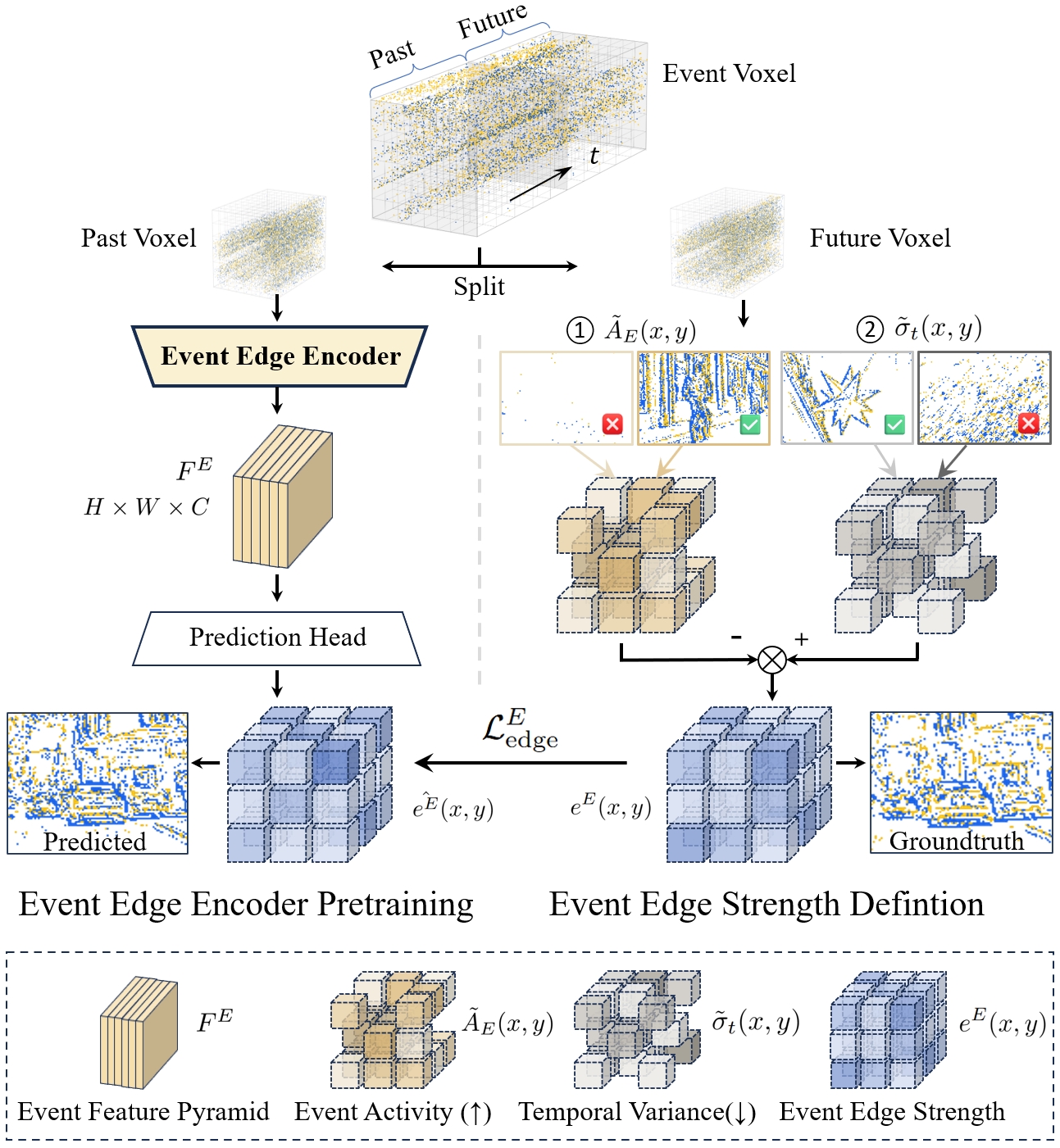}
     \caption{ Our proposed event edge encoder pretraining learns explicit, high-fidelity motion-aware edge representations by predicting edge strength from voxelized event streams.
     }
    \label{encoder}
    \vspace{-0.3cm}
\end{figure} 

We process event stream by first voxelizing it in space-time. The resulting voxel tensor is then fed into a sparse 3D convolutional neural network, which outputs a multi-scale event feature pyramid $\{F^E\}$, where $F^E_s \in \mathbb{R}^{H_s \times W_s \times C_s}$ denotes the event edge feature map at scale $s$.

To explicitly bias $F^E_s$ towards edge-aware representations, we define an event edge strength from raw events. For each pixel $(x,y)$ in a window, we measure the normalized event activity $\tilde A^E(x,y)$ and the normalized temporal variance $\tilde\sigma_t(x,y)$ of its events, and combine them into
\begin{equation}
    e^{E}(x,y) = \tilde A^E(x,y)\,\bigl(1 - \tilde\sigma_t(x,y)\bigr) \in [0,1],
\end{equation}
where larger $e^E$ indicates stronger and more temporally coherent motion edges. Scale-specific edge maps $e^E_s$ are obtained by spatial pooling of $e_E$ to the resolution $(H_s,W_s)$. The definitions of $\tilde A^E(x,y)$ and $\tilde\sigma_t(x,y)$  are in the supplementary material.

We pretrain the event encoder self-supervisedly by predicting future edge strength from past events. Specifically, for each temporal window, we split it into two halves: the first half is voxelized as the encoder input, and the latter provide the edge groundtruth. The encoder consumes the past voxels, generating $F^{E,\mathrm{past}}_s$, on top of which a lightweight prediction head $g_s$ regresses $e^{E,\mathrm{future}}_s$ with scale weights $\lambda_s$:
\begin{equation}
    \mathcal{L}_{\mathrm{edge}}^E
    = \textstyle\sum_s \lambda_s \,\bigl\|\,
        g_s\!\big(F^{E,\mathrm{past}}_s\big)
        - e^{E,\mathrm{future}}_s
      \bigr\|_1.
\end{equation}
This pretraining step distills stable, motion-aware edge features into the multi-scale event pyramid $\{F^E_s\}$, which subsequently define the event-guided prototype space used for Image–Event–LiDAR alignment in Sec.~\ref{sec:RGB--LiDAR Alignment in Event Edge Space}.

\subsubsection{Image--LiDAR Alignment in Event Edge Space}
\label{sec:RGB--LiDAR Alignment in Event Edge Space}

\paragraph{Image and LiDAR encoders.}
We employ modality-specific backbones to extract multi-scale features from images and LiDAR point clouds. The image Encoder processes image frames and derived cues to produce a 2D feature pyramid $F^I \in \mathbb{R}^{H \times W \times C^{2D}}$. The LiDAR Encoder processes consecutive point clouds, augmenting point-wise features with geometric and depth descriptors, yielding 3D point-wise feature pyramids $F^L \in \mathbb{R}^{N \times C^{3D}}$. Full architectural details and input augmentations are provided in the supplementary material.

\vspace{-0.5cm}
\paragraph{Projection.}
To map all modalities into the shared Event Edge Space, we attach lightweight projection heads ($h^I_s$, $h^L_s$) to the Image ($F^I$) and LiDAR ($F^L$) feature pyramids. These heads, implemented as a $1{\times}1$ convolution and a shared MLP respectively, project features channel-wise into the common embedding dimension $C_s$:
\begin{equation}
    Z^I_s = h^I_s(F^I_s), \quad
    Z^{L}_s = h^L_s(F^L_s), \quad
    Z^E_s \equiv F^E_s.
\end{equation}
In this homogeneous space $Z_s$, all features share the same dimension $C_s$ and are used for alignment. Crucially, the pretrained event features $Z^E_s$ are treated as fixed edge prototypes; we freeze the event encoder and stop the gradient at $Z^E_s$ during multimodal training.
\vspace{-0.5cm}
\paragraph{Edge-anchored symmetric regularization.}
We employ the frozen event embeddings $Z^E$ as stable edge prototypes to symmetrically regularize image ($Z^I$) and LiDAR ($Z^L$) features in the homogeneous space. This process effectively transfers the refined edge semantics to both the 2D (optical flow) and 3D (scene flow) task branches.

For the 2D Branch (pixel-wise optical flow), features are aligned on the image plane. We follow an interpolation strategy (based on Bi-CLFM~\cite{liu2022camliflow}) to obtain the pixel-wise LiDAR embeddings $Z^{L,2D} \in \mathbb{R}^{H \times W \times C}$ from $Z^{L}$, thus aligning $Z^{L,2D}$ with the 2D grid features $Z^I$ and $Z^E$.

For the 3D Branch (point-wise scene flow), features are aligned in the point cloud domain. We sample the image ($Z^I$) and event ($Z^E$) embeddings at the projected point locations to obtain point-wise features $Z^{I,3D}, Z^{E,3D} \in \mathbb{R}^{N \times C}$. Similarly, the event edge map $e^E$ is sampled to yield point-wise weights $e^{E,3D}(i)$.

We measure per-location alignment by the $\ell_1$ distance as

\vspace{-0.3cm}
\begin{equation}
    D_s^{2/3D}(p)
    =
    {\textstyle\sum_{(m,n)\in\{I,E,L\}}}
    \big\| Z^m_s(p) - Z^n_s(p) \big\|_1,
\end{equation}
where $p$ denotes either a 2D pixel $(x,y)$ or a 3D point $i$, yielding $D^{2D}_s(x,y)$ and $D^{3D}_s(i)$, respectively.  Note that we stop the gradient at $Z^E_s(p)$ to fix edge prototypes.

The weighted 2D and 3D alignment losses are then defined using the event edge map $e^E$ as weighting prior:
\begin{equation}
    \mathcal{L}_{\mathrm{align}}^{2/3D}
    =
    \textstyle\sum_s \sum_{p}
    e^{E/\{E,3D\}}_s(p)\, D^{2/3D}_s(p),
\end{equation}
and the total alignment loss is 
\begin{equation}
    \mathcal{L}_{\mathrm{align}} = \lambda_{2D}\cdot\mathcal{L}_{\mathrm{align}}^{2D} + \lambda_{3D}\cdot\mathcal{L}_{\mathrm{align}}^{3D}.
\end{equation}

\subsection{Reliability-aware Adaptive Fusion}
\label{sec:event_aware_adaption_fusion}
Benefiting from the tri-modal alignment in the Event Edge Space, we fuse features directly within this shared latent domain, eliminating the need for bespoke modality-specific interaction modules. We leverage event data's strengths, temporal fine-grained information and high dynamic range, to achieve robust adaptive fusion. Our method employs a symmetric dual-branch decoder (2D/3D), where each branch utilizes an reliability-aware adaptive fusion module to dynamically weight all three modalities based on sensor reliability for robust fusion.

\vspace{-0.5cm}
\noindent\paragraph{Adaptive fusion.}
Leveraging the homogeneous embeddings $\{Z^I_s, Z^L_s, Z^E_s\}$, we utilize two lightweight feature decomposition streams: a motion stream aggregating features along temporal dimension to capture global motion consistency, and a spatial stream aggregating features within local neighborhoods to capture local structural agreement.

To facilitate adaptive integration, we estimate a global reliability score $\omega_m$ for each modality $m \in \{\text{I}, \text{L}\}$ by measuring its consistency with the event-derived motion signal. Concretely, this score is derived via a spatiotemporal decomposition applied to the concatenated feature map $\hat{Z} = [Z_M, Z_E]$:
\begin{equation}
\mathcal{T}(\hat{Z}) = \sigma ( \mathbb{L} ( \Delta_t ( \text{Conv}(\hat{Z}) ))), 
\mathcal{S}(\hat{Z}) = \| \nabla ( \text{DConv}(\hat{Z}) ) \|_2.
\end{equation}
Here, $\mathcal{T}(\cdot)$ captures fine-grained temporal changes using feature differencing $\Delta_t$, followed by convolution and a linear projection layer $\mathbb{L}$; $\mathcal{S}(\cdot)$ encodes spatial structure via dilated convolutions and gradient magnitude.
The resultant global reliability score is computed as:
\begin{equation}
\omega_m = \text{softmax}_m ((\mathcal{T}\otimes \mathcal{S})\hat{Z}).
\end{equation}

\begin{figure}[t]
    \centering
        \includegraphics[width=1.0\linewidth]{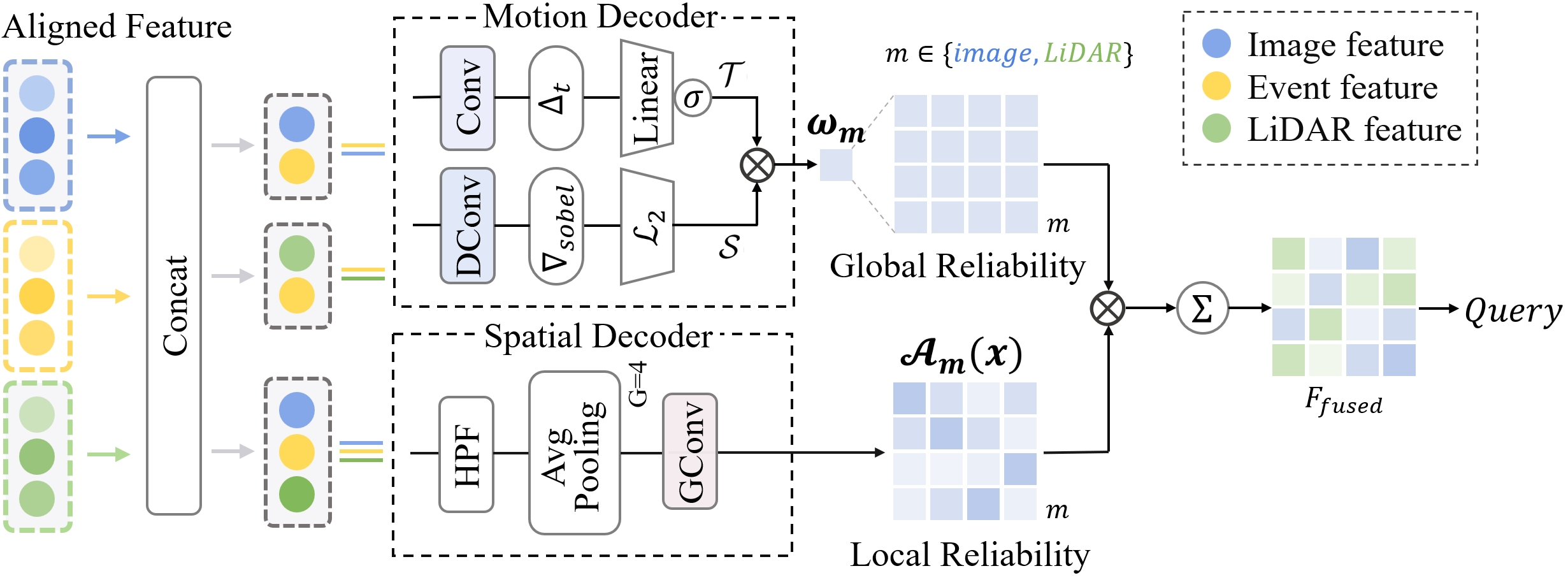}
     \caption{ The proposed reliability-aware adaptive fusion module adaptively integrates image, LiDAR, and event features through hierarchical reliability weighting and cross-modal attention.}
    \label{bridge2}
    \vspace{-0.3cm}
\end{figure}
To further account for local spatial variations in reliability, we design a modality-specific attention mechanism that processes the tri-modal feature volume $\tilde{Z} = [Z_{I}, Z_{{L}}, Z_{E}]$ via a lightweight network:
\begin{equation}
\mathcal{A}_m(x) = \text{softmax}((\mathcal{H} \oplus \mathcal{P} \oplus \mathcal{G})\tilde{Z})_m,
\end{equation}
where $\mathcal{H}$, $\mathcal{P}$, and $\mathcal{G}$ denote high-pass filtering (HPF), average pooling, and grouped $1{\times}1$ convolutions, respectively. The output $\mathcal{A}_m(x)$ reflects the spatially-varying reliability of modality $m$ at location $x$.

Finally, the fused feature $F_{{fused}}(x)$ is computed as a weighted sum over modalities. By combining the global reliability score and local attention, this mechanism adaptively emphasizes key modalities at both scales to enhance reliability of the model.
\begin{equation}
F_{{fused}}(x) = \sum_{m\in\mathcal{M}} \frac{\omega_m\mathcal{A}_m(x)}{\sum_n \omega_n\mathcal{A}_n(x)} Z_m(x).
\end{equation}

\vspace{-0.5cm}
\noindent\paragraph{Cross-attention transformer.}
To enhance multimodal interaction, we introduce a cross-attention transformer for flow estimation. 
For 2D and 3D branch, the module takes fused features $F_{fused}$ as queries and projects auxiliary features $[Z^{L,2D}, Z^E]$ and $[Z^{I,3D}, Z^{E,3D}]$ via learnable MLPs.

The cross-attention is implemented using a transformer block, where queries, keys, and values are computed as:
\begin{equation}
Q = W_q F_{fused}, \quad K = W_k F_{aux}, \quad V = W_v F_{aux}.
\end{equation}
Here, $W_q$, $W_k$, and $W_v$ are implemented as $1{\times}1$ convolutions for 2D and linear layers for 3D, enabling modality- and domain-specific adaptation.
The attention output is given by:
\begin{equation}
\text{Attention}(Q, K, V) = V \cdot \text{softmax}\left(\frac{Q K^\top}{\sqrt{d}}\right),
\end{equation}
where $d$ is the feature dimension. The attended features are refined via a feed-forward MLP and combined with the input using residual connections:
\begin{equation}
F_{{out}}^{2/3D} = \text{MLP}(\text{Attention}(Q, K, V)^{2/3D}) + F_{fused}.
\end{equation}

\begin{table*}[t]
\scriptsize
\renewcommand{\arraystretch}{1.05} 
\setlength{\tabcolsep}{1.91pt}        
\vspace{-0.3cm}
\centering 
\begin{tabular}{@{}c*{15}{c}@{}}
\toprule
\multirow{2}{*}{\centering Modality} & \multirow{2}{*}{\centering Method} & \multirow{2}{*}{\centering \#Params(M)} & \multicolumn{6}{c}{EKubric} & \multicolumn{6}{c}{DSEC} \\
\cmidrule(r){4-9}\cmidrule(l){10-15}
& & & $\text{EPE}_\text{2D}$ $\downarrow$ & $\text{ACC}_\text{1px}$ $\uparrow$ & $\text{Fl}$ $\downarrow$ &  $\text{EPE}_\text{3D}$ $\downarrow$ & $\text{ACC}_\text{.05}$ $\uparrow$ & $\text{ACC}_\text{.10}$ $\uparrow$ & $\text{EPE}_\text{2D}$ $\downarrow$ & $\text{ACC}_\text{1px}$ $\uparrow$ & $\text{Fl}$ $\downarrow$ & $\text{EPE}_\text{3D}$ $\downarrow$ & $\text{ACC}_\text{.05}$ $\uparrow$ & $\text{ACC}_\text{.10}$ $\uparrow$ \\


\cmidrule{1-15}


Img & RAFT~\cite{teed2020raft} & 5.3M & 0.838 & 93.31\% & 2.36\% & - & - & - & 0.586 & 88.98\%  & 1.47\% & - & - & - \\ 
Img & FlowFormer~\cite{huang2022flowformer} & 16.2M & 0.702 & 92.58\% & 2.07\% & - & - & - & 0.567 & 89.82\% & 1.33\% & - & - & - \\ 
PC{\tiny (Point Cloud)} & PV-RAFT~\cite{wei2021pv} & 239.2K & - & - & - & 0.093 & 82.42\% & 92.60\% & - & - & - & 0.190 & 32.74\% & 55.62\% \\ 
EV & E-RAFT~\cite{gehrig2021raft} & 5.3M & - & - & - & - & - & - & 0.481 & 91.75\% & 1.31\% & - & - & - \\
Img+Depth & RAFT-3D~\cite{teed2021raft} & 7.7M & 0.714 & 94.39\% & - & 0.049 & 92.88\% & - & 0.572 & 89.55\%  & - & 0.144 & 49.30\% & - \\
Img+PC & CamLiFlow~\cite{liu2022camliflow} & 7.7M & 0.770 & 95.11\% & 1.80\% & 0.035 & 94.90\% & 95.86\% & 0.399 & 94.94\%  & 1.33\% & 0.129 & 51.08\% & 68.17\% \\
Img+EV & Diff-ABFlow~\cite{wang2025injecting} & 17.2M & - & - & - & - & - & - & 1.460 & 50.00\% & 7.43\% & - & - & - \\
Img+EV & DCEIFlow~\cite{wan2022DCEIFlow} & 7.1M & 3.109 & 56.37\% & 18.40\% & - & - & - & 0.970 & 73.01\% & 4.90\% & - & - & - \\
Img+EV & STFlow~\cite{zhou2026spatially} & 9.2M & - & - & - & - & - & - & 0.630 & 92.07\%  & 1.45\% & - & - & - \\
Img+PC+EV & $^\dagger$VisMoFlow~\cite{zhou2024bring} & - & - & - & - & \underline{0.026} & \underline{95.98\%} & \underline{96.77\%} & - & - & - & \underline{0.100}  & \underline{61.78\%} & \underline{75.82\%} \\
        Img+PC+EV & RPEFlow~\cite{wan2023rpeflow} & 9.8M & \underline{0.439} & \underline{95.99\%} & \underline{1.48\%} & 0.027 & 95.33\% & 96.32\% & \underline{0.326} & \underline{95.28\%}  & \underline{1.15\%} & 0.103 & 60.80\% & 74.97\% \\ 
\rowcolor{blue!7}
Img+PC+EV & $\text{\model}$ (Ours) & 8.2M & \textbf{0.430} & \textbf{96.86\%} & \textbf{1.43\%} & \textbf{0.024} & \textbf{96.78\%} & \textbf{97.62\%} & \textbf{0.322} & \textbf{95.80\%} & \textbf{1.13\%} & \textbf{0.094} & \textbf{64.39\%}  & \textbf{78.27\%} \\

\cmidrule{1-15}
\rowcolor{gray!10}
\multicolumn{3}{c}{ } & \multicolumn{6}{c}{EKubric-Img (Under-Exposure Degradation)} & \multicolumn{6}{c}{DSEC-Img (Under-Exposure Degradation)} \\
\cmidrule(r){4-9}\cmidrule(l){10-15}

Img+PC+EV & RPEFlow~\cite{wan2023rpeflow}  & & 3.663 & 33.96\% & 27.06\% & 0.043 & 89.01\% & 92.27\% & 0.817 & 83.24\%  & 5.06\% & 0.117 & 54.97\% & 69.94\% \\
\rowcolor{blue!7}
Img+PC+EV &  \ $\text{\model}$ (Ours)  & &  \textbf{1.143} & \textbf{68.87\%} & \textbf{9.11\%} & \textbf{0.033} & \textbf{94.14\% } & \textbf{95.31\%}  & \textbf{0.394} & \textbf{92.73\%} & \textbf{1.14\%} & \textbf{0.107}  & \textbf{58.87\%} & \textbf{71.43\%} \\
\multicolumn{3}{c}{\textit{Absolute Improvement (vs. RPEFlow)}} & {\tiny \blue{($\downarrow$ 2.520)}} & {\tiny \blue{($\uparrow$ 34.91\%)}} & {\tiny \blue{($\downarrow$ 17.95\%)}} & {\tiny \blue{($\downarrow$ 0.010)}} & {\tiny \blue{($\uparrow$ 5.13\%)}}  & {\tiny \blue{($\uparrow$ 3.04\%)}}  & {\tiny \blue{($\downarrow$ 0.423)}} & {\tiny \blue{($\uparrow$ 9.49\%)}} & {\tiny \blue{($\downarrow$ 3.92\%)}} & {\tiny \blue{($\downarrow$ 0.010)}}  &  {\tiny \blue{($\uparrow$ 3.90\%)}} & {\tiny \blue{($\uparrow$ 1.49\%)}} \\

\cmidrule{1-15}
\rowcolor{gray!10}
\multicolumn{3}{c}{ }  & \multicolumn{6}{c}{EKubric-Img (Over-Exposure Degradation)} & \multicolumn{6}{c}{DSEC-Img (Over-Exposure Degradation)} \\
\cmidrule(r){4-9}\cmidrule(l){10-15}

Img+PC+EV & RPEFlow~\cite{wan2023rpeflow}  &  & 2.801 & 53.75\% & 17.63\% & 0.039 & 91.32\% & 95.85\% & 0.565 & 89.59\%  & 2.73\% & 0.109 & 54.91\% & 70.02\% \\
\rowcolor{blue!7}
Img+PC+EV &  $\text{\model}$ (Ours) &   & \textbf{0.994} & \textbf{80.19\%} & \textbf{7.05\%} & \textbf{0.032} & \textbf{93.86\%}  & \textbf{95.93\%}  & \textbf{0.376} & \textbf{93.03\%} & \textbf{1.13\%} & \textbf{0.107}  & \textbf{58.67\%} & \textbf{71.09\%} \\
\multicolumn{3}{c}{\textit{Absolute Improvement (vs. RPEFlow)}} & {\tiny \blue{($\downarrow$ 1.807)}} & {\tiny \blue{($\uparrow$ 26.44\%)}} & {\tiny \blue{($\downarrow$ 10.58\%)}} & {\tiny \blue{($\downarrow$ 0.007)}} & {\tiny \blue{($\uparrow$ 2.54\%)}} & {\tiny \blue{($\uparrow$ 0.08\%)}}  & {\tiny \blue{($\downarrow$ 0.189)}} & {\tiny \blue{($\uparrow$ 3.44\%)}} & {\tiny \blue{($\downarrow$ 1.6\%)}} & {\tiny \blue{($\downarrow$ 0.002)}} & {\tiny \blue{($\uparrow$ 3.76\%)}} & {\tiny \blue{($\uparrow$ 1.07\%)}} \\

\cmidrule{1-15}
\rowcolor{gray!10}
\multicolumn{3}{c}{ } & \multicolumn{6}{c}{EKubric-PC (Sparse Degradation)} & \multicolumn{6}{c}{DSEC-PC (Sparse Degradation)} \\
\cmidrule(r){4-9}\cmidrule(l){10-15}

Img+PC+EV & RPEFlow~\cite{wan2023rpeflow}  & & 0.569 & 94.62\% & 2.02\% & 0.027 & 95.37\% & 96.34\% & 0.493 & 90.36\% & 1.19\% & 0.056 & 81.18\% & 88.69\% \\
\rowcolor{blue!7}
Img+PC+EV &  \ $\text{\model}$ (Ours)  & &  \textbf{0.439} & \textbf{95.93\%} & \textbf{1.49\%} & \textbf{0.022} & \textbf{96.99\%}  & \textbf{97.43\%}  & \textbf{0.331} & \textbf{94.49\%} & \textbf{1.18\%} & \textbf{0.047 }& \textbf{86.40\%} & \textbf{90.31\%} \\
\multicolumn{3}{c}{\textit{Absolute Improvement (vs. RPEFlow)}} & {\tiny \blue{($\downarrow$ 0.130)}} & {\tiny \blue{($\uparrow$ 1.31\%)}} & {\tiny \blue{($\downarrow$ 0.53\%)}} & {\tiny \blue{($\downarrow$ 0.005)}} & {\tiny \blue{($\uparrow$ 1.62\%)}}  & {\tiny \blue{($\uparrow$ 1.09\%)}}  & {\tiny \blue{($\downarrow$ 0.162)}} & {\tiny \blue{($\uparrow$ 4.13\%)}} & {\tiny \blue{($\downarrow$ 0.01\%)}} & {\tiny \blue{($\downarrow$ 0.009)}} & {\tiny \blue{($\uparrow$ 5.22\%)}} & {\tiny \blue{($\uparrow$ 1.62\%)}} \\

\cmidrule{1-15}
\rowcolor{gray!10}
\multicolumn{3}{c}{ }  & \multicolumn{6}{c}{EKubric-PC (Drifting Degradation)} & \multicolumn{6}{c}{DSEC-PC (Drifting Degradation)} \\
\cmidrule(r){4-9}\cmidrule(l){10-15}

Img+PC+EV & RPEFlow~\cite{wan2023rpeflow}  & & 1.164 & 62.44\% & 9.92\% & 0.113 & 66.47\% & 73.76\% & 1.051 & 68.05\% & 3.82\% & 0.457 & 0.51\% & 2.46\% \\
\rowcolor{blue!7}
Img+PC+EV &  \ $\text{\model}$ (Ours)  & & \textbf{0.763} & \textbf{88.52\%} & \textbf{6.06\%} & \textbf{0.088} & \textbf{79.34\%} & \textbf{82.80\%} & \textbf{0.409} & \textbf{91.39\%} & \textbf{2.13\%} & \textbf{0.285} & \textbf{19.36\%} & \textbf{28.87\%} \\
\multicolumn{3}{c}{\textit{Absolute Improvement (vs. RPEFlow)}} & {\tiny \blue{($\downarrow$ 0.401)}} & {\tiny \blue{($\uparrow$ 26.08\%)}} & {\tiny \blue{($\downarrow$ 3.86\%)}} & {\tiny \blue{($\downarrow$ 0.025)}} & {\tiny \blue{($\uparrow$ 12.77\%)}} & {\tiny \blue{($\uparrow$ 9.04\%)}} & {\tiny \blue{($\downarrow$ 0.642)}} & {\tiny \blue{($\uparrow$ 23.34\%)}} & {\tiny \blue{($\downarrow$ 1.69\%)}} & {\tiny \blue{($\downarrow$ 0.172)}} & {\tiny \blue{($\uparrow$ 18.85\%)}} & {\tiny \blue{($\uparrow$ 26.41\%)}} \\

\bottomrule
\end{tabular}
\vspace{-0.1cm}
\caption{Quantitative evaluation on both standard and degraded scenarios of EKubric and DSEC datasets. 
\textsuperscript{\scriptsize$\dagger$} Reproduced per original methodology (code unavailable).
Extended results are provided in the supplementary material.
}
\label{tab:accuracy}
\end{table*}

\subsection{Cross-dimension Contrast Learning (CCL)}
\label{sec:CCL}
Inspired by mutual information optimization in multimodal fusion, we extend to cross-temporal and cross-task constraints. Specifically, we regularize the 2D and 3D features from cross-attention module, enhancing their intra-frame distinctiveness and inter-frame consistency.

\noindent\textbf{Cross-temporal contrast (pull).}
To capture inter-frame dynamics, we compute temporal differences of 2D and 3D fused features. As 3D features lie in a different spatial domain, we project them to 2D space to obtain ${F}^{3D}_{proj}$. Temporal motion vectors ${M}^{2D}$ and ${M}^{3D}$ are then computed via global average pooling:
\begin{equation}
M^{2/3D} = \mathcal{P}(\Delta F_{out}^{2/3D}).
\end{equation}
A cosine similarity loss encourages geometric alignment between normalized motion embeddings:
\begin{equation}
\mathcal{L}_{pull} = 1 - \frac{\langle \phi(M^{2D}), \psi(M^{3D}_{proj}) \rangle}{\|\phi(M^{2D})\|_2 \cdot \|\psi(M^{3D}_{proj})\|_2},
\end{equation}
where $\phi(\cdot)$ and $\psi(\cdot)$ denote normalization functions.

\noindent\textbf{Cross-task contrast (push).}
To encourage complementary intra-frame representations, we minimize the mutual information between 2D and 3D features via variational encoding. Each modality is encoded into a latent distribution:
\begin{equation}
\begin{aligned}
        q_\phi(\mathbf{z}^{2D}|{F}^{2D})=& \mathcal{N}(\boldsymbol{\mu}^{2D}, \boldsymbol{\sigma}^{2D}), \\q_\psi(\mathbf{z}^{3D}|{F}^{3D}_{proj})=& \mathcal{N}(\boldsymbol{\mu}^{3D}, \boldsymbol{\sigma}^{3D}),
\end{aligned}
\end{equation}
where $\mathbf{z}^{2D}$ and $\mathbf{z}^{3D}$ represent the latent embeddings sampled from these distributions.

The mutual information is approximated by a symmetric binary cross-entropy(BCE) loss over sampled latent pairs:
\begin{equation} 
 \mathcal{L}_{{push}} = \frac{1}{2} \textstyle\sum_{t \in \{t_1, t_2\}} \text{BCE}(\sigma(\mathbf{z}^{2D}_{t}), \sigma(\mathbf{z}^{3D}_{t})).
\end{equation} 
The total contrastive loss is a weighted sum of the above:
\begin{equation} 
 \mathcal{L}_{{contra}} = \mathcal{L}_{{pull}} + \gamma \cdot \mathcal{L}_{{push}}.
\end{equation} 

\subsection{Total Objective Function}\label{sec:total_loss}
Our training objective is a weighted sum of three losses: flow supervision, feature stabilization, and cross-modal alignment.
\begin{equation}
\mathcal{L}_{\mathrm{total}} = \mathcal{L}_{\mathrm{task}} + \lambda_{\mathrm{align}} \mathcal{L}{_\mathrm{align}} + \lambda_{\mathrm{contra}} \mathcal{L}_{\mathrm{contra}}.
\end{equation}

\noindent\textbf{Task loss} ($\mathcal{L}_{\mathrm{task}}$).
For joint flow estimation, we adopt PWC-style coarse-to-fine supervision, upsampling and warping the level-$l$ estimate to initialize level $l{-}1$.
\begin{equation}
\setlength\abovedisplayskip{3pt}
\setlength\belowdisplayskip{3pt}
\mathcal{L}_{\mathrm{task}}
= \textstyle\sum_{2/3D} \lambda_{2/3D} \sum_{l} \omega^{2/3D}_l
\left\| \mathbf{f}^{2/3D}_{\mathrm{pred},l} - \mathbf{f}^{2/3D}_{\mathrm{gt},l} \right\|_2 .
\end{equation}


\noindent\textbf{Alignment loss} ($\mathcal{L}_{\mathrm{align}}$). Defined in Sec.~\ref{sec:RGB--LiDAR Alignment in Event Edge Space}, this loss ensures Image and LiDAR features are consistently aligned to the fixed event prototypes in the homogeneous space.

\noindent\textbf{Contrastive loss} ($\mathcal{L}_{\mathrm{contra}}$). Introduced in Sec.~\ref{sec:CCL}, this auxiliary term regularizes the 2D and 3D features, enhancing their inter-frame consistency ($\mathcal{L}_{\mathrm{pull}}$) and intra-frame complementarity ($\mathcal{L}_{\mathrm{push}}$).

\noindent The event encoder is initially optimized using the self-supervised edge loss $\mathcal{L}_{\mathrm{edge}}^E$ (Sec.~\ref{sec:event_edge_encoder}) and is subsequently frozen during the optimization of $\mathcal{L}_{\mathrm{total}}$.

\begin{figure*}[h]
    \centering
        \includegraphics[width=1.0\linewidth]{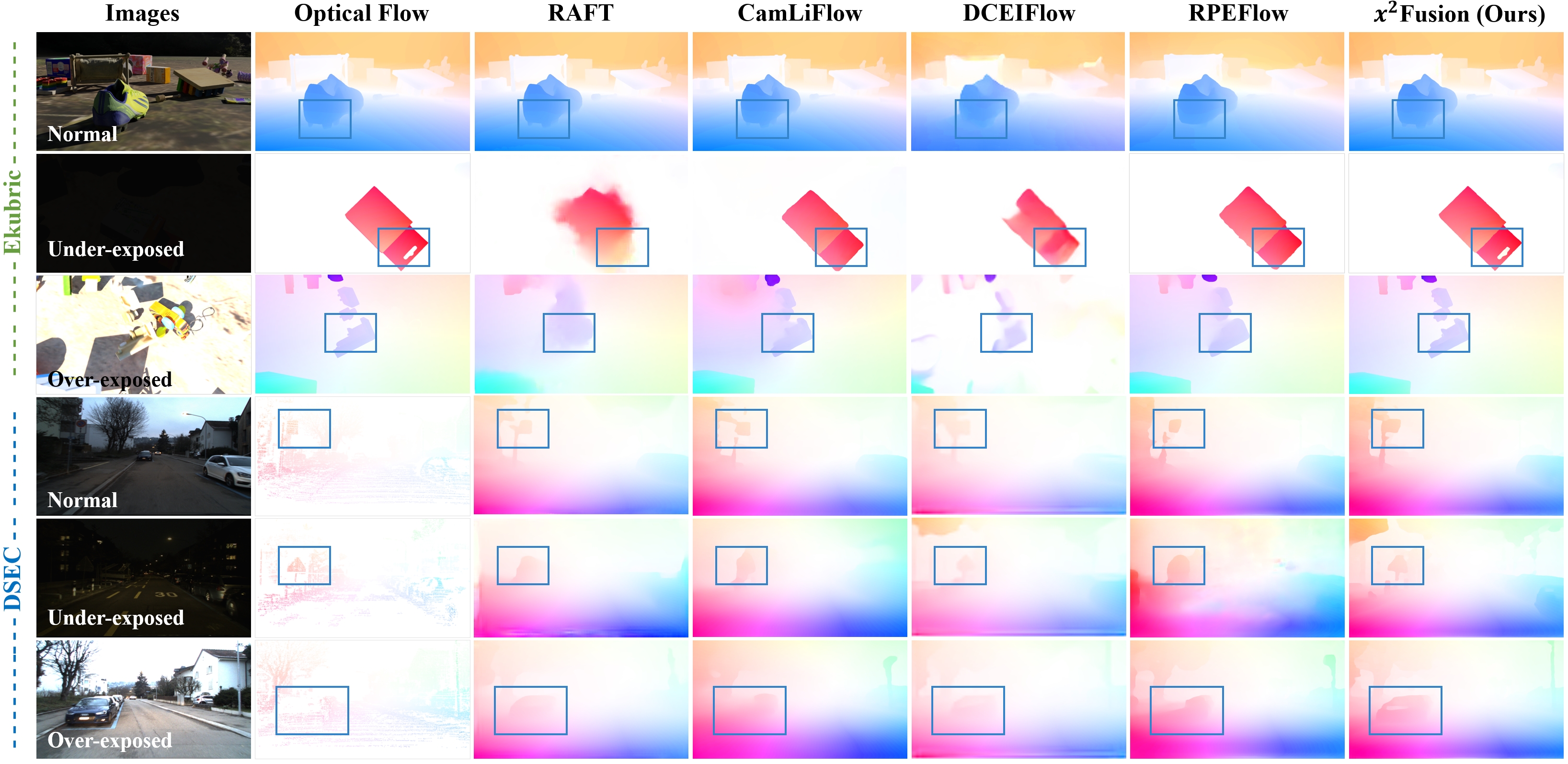}
    \vspace{-0.5cm}
     \caption{ Visual comparison of optical flows on EKubric and DSEC dataset. \model \ achieves the state-of-the-art performance across various exposure degradation scenarios with clearer motion boundaries and finer details.
     Please zoom in for details.
     }
    \label{2D}
    \vspace{-0.1cm}
\end{figure*} 

\begin{figure*}[h]
    \centering
        \includegraphics[width=1.0\linewidth]{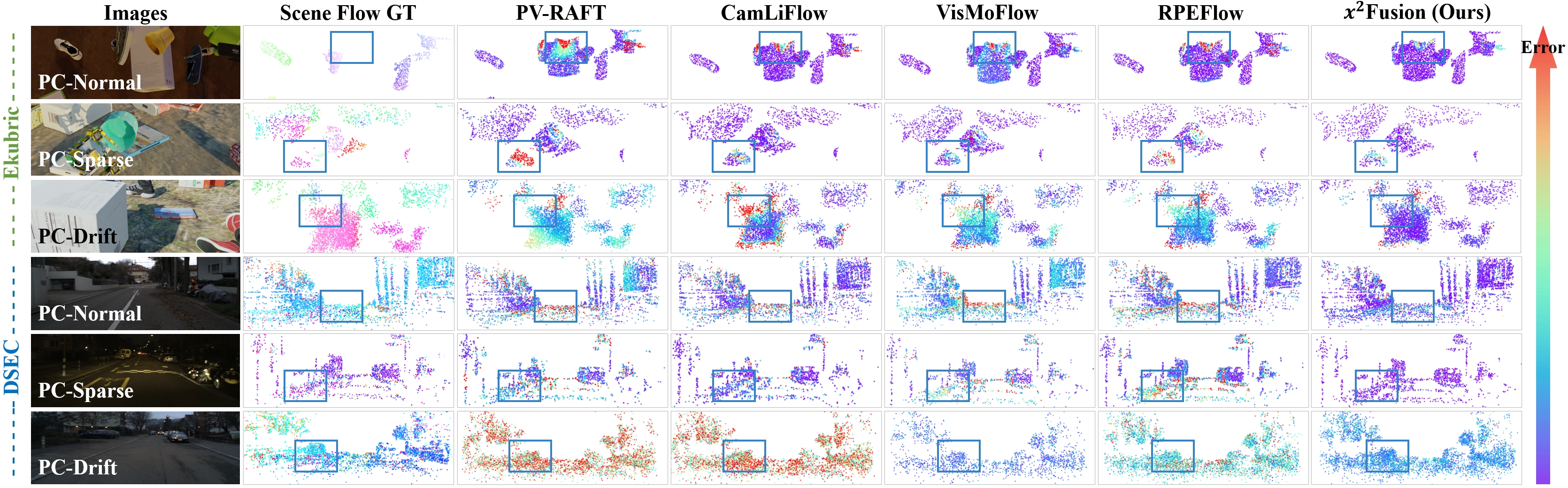}
    \vspace{-0.5cm}
     \caption{ Visual comparison of scene flows on EKubric and DSEC dataset. \model \ demonstrates 
     achieves lower end-point errors across diverse degradation scenarios.
     Comprehensive results and analyses are provided in the supplementary material.
     Please zoom in for details.
     }
    \label{3D}
    \vspace{-0.3cm}
\end{figure*}

\section{Evaluation}
\subsection{Experiments Setups}
\paragraph{Datasets and metrics.}
We conduct extensive experiments on both \textbf{\textit{synthetic}} (EKubric) and \textbf{\textit{real-world}} (DSEC) dataset. 
EKubric provides 15,367 annotated Image-LiDAR-Event triplets, while DSEC captures diverse urban driving scenarios with complex traffic elements.  
To stress-test robustness under realistic extreme conditions, we synthesize four degradations (low/high-exposure image and sparse/drifting LiDAR) using empirically grounded noise models\cite{liang2024towards, dong2023benchmarking}, producing artifacts that mirror common degradation modes.
For quantitative evaluation, we employ $\text{EPE}_\text{2D}$ (End-Point Error), $\text{ACC}_\text{1px}$ (accuracy within 1 pixel), and Fl (outlier rate with $\text{EPE}_\text{2D}$$>3$px \& error$>$5\%) for 2D Flow estimation; and $\text{EPE}_\text{3D}$, and $\text{ACC}_{.05}$/$\text{ACC}_{.10}$ (accuracy within 5/10 cm) for 3D Flow estimation.

\vspace{-0.5cm}
\paragraph{Implementation details.}
Our model is implemented in PyTorch, trained on four NVIDIA RTX A6000 GPUs and evaluated on a single GPU. We use the Adam optimizer with an initial learning rate of $10^{-4}$, weight decay of $10^{-6}$, and batch size of 8, using MultiStepLR scheduling with 0.5$\times$ decay at specified milestones. 
All experiments leverage synchronized batch normalization and mixed-precision training for efficient optimization.

\subsection{Comparison Experiment}


In Tab.~\ref{tab:accuracy} and Fig.~\ref{2D}–\ref{3D}, we evaluate methods on synthetic and real-world datasets under normal and degraded conditions (e.g., extreme exposure, LiDAR corruption). 
First, \model \ achieves state-of-the-art optical and scene flow estimation, consistently improving EPE and accuracy metrics over recent baselines.
Second, it remains highly robust against various degradations (including under/over-exposure and sparse/drifting point clouds), effectively mitigating partial data failure. 
Third, unified tri-modal fusion surpasses bi/uni-modal variants, with the largest gains in degraded conditions, proving that complementary modalities provide more stable and reliable motion references.

\subsection{Ablation Study}
\paragraph{How does Event Edge Space work?}
In Tab.~\ref{tab:abaltion_1} and Fig.~\ref{bridge2}, we demonstrate the effectiveness of the proposed Event Edge Space.
All variants are trained on the DSEC training split and evaluated on its validation split. 
When the Event Edge Space (EES) is removed(w/o EES), cross-attention must simultaneously repair heterogeneous feature geometries and model motion. As shown in Tab.~\ref{tab:abaltion_1}, performance drops on both optical and scene flow, indicating that an explicit edge-centric latent space offloads alignment and allows the decoder to focus on motion modeling over already-aligned embeddings.
When the symmetric regularization is ablated(w/o Reg), accuracy improves over w/o EES yet remains below the full model; t-SNE in Fig.~\ref{bridge2} shows that the regularizer pulls image/LiDAR embeddings toward event-defined clusters, forming tighter cross-modal neighborhoods and providing better-aligned inputs. 
When the event edge encoder is left unpretrained(w/o Edge), results lie between w/o EES and the full model, suggesting that an event pathway already helps unify the space, while reliability-aware edge pretraining is crucial for a meaningful prototype and for transferring edge semantics. 
Overall, edge prototypes provide a fixed alignment anchor and symmetric regularization pulls image/LiDAR toward it, producing better-aligned inputs.

\begin{table}[t]
\footnotesize
    \centering
    \setlength{\tabcolsep}{4pt} 
     \renewcommand{\arraystretch}{1.0}
    \begin{tabular}{ccccc}
        \toprule
         Settings &$\text{EPE}_\text{2D}$$\downarrow$ & 
        $\text{EPE}_\text{3D}$$\downarrow$  \\
       \cmidrule{1-3}
        w/o Event Edge Space    & 0.491 & 0.146   \\
        w/o Edge-anchored Regularization    & 0.393 & \underline{0.119}    \\
        w/o Event Edge Encoder   & 0.378 & 0.114 \\
        w/ Event Edge Space     & \textbf{0.322} & \textbf{0.094} \\
        \bottomrule
    \end{tabular}
    \caption{Effects of the Event Edge Space.}
    \label{tab:abaltion_1}
    \vspace{-0.3cm}
\end{table}

\begin{figure}[t]
    \centering
        \includegraphics[width=1.0\linewidth]{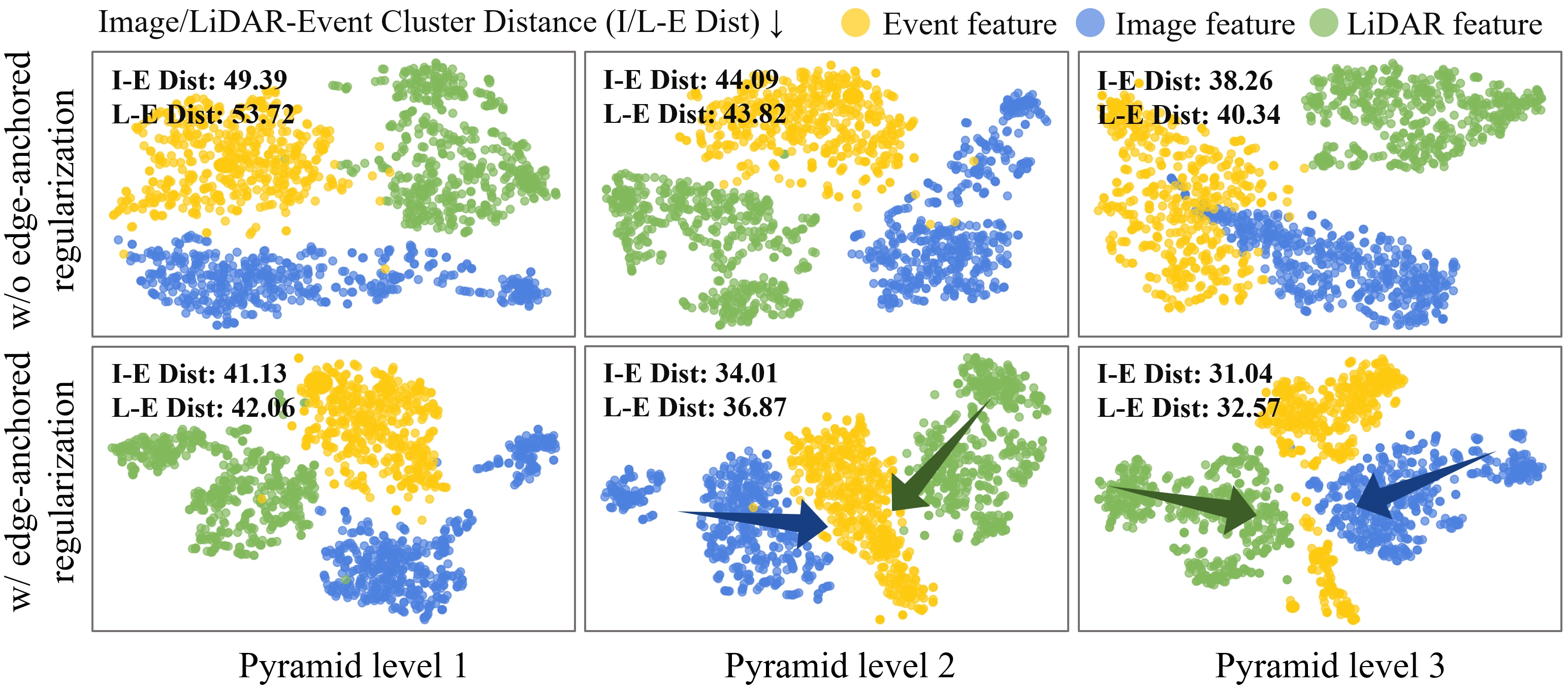}
     \caption{ t-SNE visualization of cross-modal feature alignment.
The edge-anchored symmetric regularization reduces tri-model feature divergence through pyramid-level alignment.}
    \label{bridge2}
    \vspace{-0.3cm}
\end{figure}

\begin{figure}[h]
    \centering
        \includegraphics[width=1.0\linewidth]{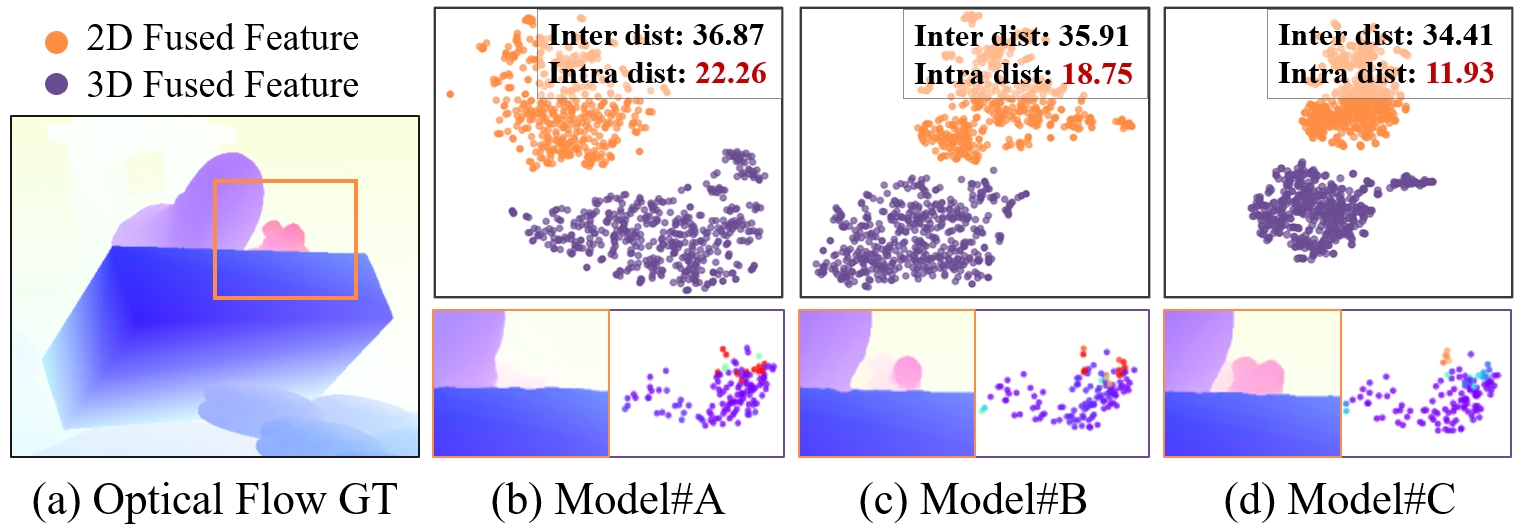}
     \caption{Qualitative analysis of joint task performance and contrastive learning effects. (Top) t-SNE visualization of fused 2D-3D feature distributions; (Bottom) Visualization of (left) 2D optical flow and (right) 3D scene flow error maps.}
    \label{ablation_3}
    \vspace{-0.5cm}
\end{figure}

\vspace{-0.3cm}
\paragraph{Why does cross-dimension learning and joint task outperform independent tasks?}
To further assess the necessity of cross-dimensional learning and joint task estimation, we compare models trained with separate supervision for optical and scene flow (Model\#A) in Tab.~\ref{tab:comparison} and Fig.~\ref{ablation_3}.
Although all models incorporate three-modal inputs, the joint estimation variant (Model\#B) achieves better performance by leveraging the inherent correlations between 2D and 3D branches.
Building upon this, our model integrates early-stage event-guided fusion, which facilitates cross-dimensional interactions (Model\#C) while boosting estimation accuracy.
The t-SNE visualization in Fig.~\ref{ablation_3} reveals that our framework enhances feature representation in two complementary ways: it yields more compact intra-class distributions to improve task-specific discriminability, while preserving inter-class relationships that support consistent scene understanding across dimensions.
These improvements are further supported by optical and scene flow results, where reduced error demonstrate the benefits of unified estimation and cross-dimensional learning.

\begin{table}[t]
\small
    \centering
    \setlength{\tabcolsep}{8pt} 
     \renewcommand{\arraystretch}{1.0}
    \begin{tabular}{clcc}
        \toprule
        Variants & Setting & $\text{EPE}_\text{2D}$$\downarrow$ & $\text{EPE}_\text{3D}$$\downarrow$ \\
        \cmidrule{1-4}
        Model\#A & Indep. 2D + 3D  & 0.404 & 0.119 \\
        Model\#B & Joint 2D \& 3D & \underline{0.386} & \underline{0.113}  \\
        Model\#C & Joint 2D \& 3D + CCL & \textbf{0.325} & \textbf{0.103} \\
        \bottomrule
    \end{tabular}
    \caption{Effects of cross-dimension learning and joint task.}
    \label{tab:comparison}
    \vspace{-0.5cm}
\end{table}

\section{Conclusion}

We introduce \textbf{\textit{Event Edge Space(EES)}}, the first edge-centric homogeneous latent space that unifies images, LiDAR, and events in a common feature domain. 
Building on it, we present \textbf{\textit{\model \ }}, which reframes multimodal fusion as representation unification. 
Within this shared edge-centric space, \model \ couples \emph{reliability-aware adaptive fusion} and \emph{cross-dimension contrastive learning}, promoting task-specific discrimination and enforcing 2D–3D consistency. 
On both synthetic and real-world benchmarks, our method achieves state-of-the-art flow accuracy and remains robust under challenging conditions.
Beyond flow estimation, this framework offers broader insights for tasks such as text–image–video fusion and general cross-domain feature alignment, where bridging heterogeneous data is critical.

Although our study focuses on the image–LiDAR–event setting for flow estimation, EES is modality-agnostic. In principle, any modality with stable edge cues can be unified in it. In future work, we will extend it to additional sensors and validate generality beyond image–LiDAR–event.

\paragraph{Acknowledgments}

This paper was supported by the Natural Science Foundation of China under Grant 62371269, Shenzhen Low-Altitude Airspace Strategic Program Portfolio (Grant No. Z25306110), Shenzhen Science and Technology Program (Grant No. ZDCYKCX20250901094203005 and No. ZDCY202517012) and Meituan Academy of Robotics Shenzhen.
{
    \small
    \bibliographystyle{unsrt}
    \bibliography{main}
}


\end{document}